\documentclass[journal]{IEEEtran}
\usepackage{cite}
\usepackage{caption}
\usepackage{subcaption}
\usepackage{amsmath,amssymb,amsfonts}
\usepackage{mathtools}
\usepackage{algorithm,algorithmic}
\usepackage{graphicx,nicefrac}
\usepackage{textcomp}
\usepackage{xcolor}
\usepackage{bbding}
\usepackage{chngpage}
\usepackage{mathrsfs}
\usepackage{url}
\usepackage{graphicx}
\usepackage{array}
\usepackage{comment}
\usepackage[pagebackref=true,breaklinks=true,letterpaper=true,colorlinks,bookmarks=false]{hyperref}

\begin{document}

\title{An embedded deep learning system for augmented reality in firefighting applications}

\author{\IEEEauthorblockN{\textbf{Manish Bhattarai}\IEEEauthorrefmark{1}\IEEEauthorrefmark{2}, 
           \textbf{Aura~Rose~Jensen-Curtis}\IEEEauthorrefmark{1}\IEEEauthorrefmark{3},
           \textbf{Manel MartíNez-Ramón}\IEEEauthorrefmark{1}} \\
    \IEEEauthorblockA{\IEEEauthorrefmark{1} University of New Mexico, Albuquerque, NM, 87106, USA, \\
    \IEEEauthorrefmark{2} Los Alamos National Laboratory, Los Alamos, NM, 87545, USA, \\
    \IEEEauthorrefmark{3} Sandia National Labs, Albuquerque, NM, 87185, USA 
}
}




\maketitle

\begin{abstract}
Firefighting is a dynamic activity, in which numerous operations occur simultaneously. Maintaining situational awareness (i.e., knowledge of current conditions and activities at the scene) is critical to the accurate decision-making necessary for the safe and successful navigation of a fire environment by firefighters. Conversely, the disorientation caused by hazards such as smoke and extreme heat can lead to injury or even fatality. This research implements recent advancements in technology such as deep learning, point cloud and thermal imaging, and augmented reality platforms to improve a firefighter's situational awareness and scene navigation through improved interpretation of that scene. We have designed and built a prototype embedded system that can leverage data streamed from cameras built into a firefighter's personal protective equipment (PPE) to capture thermal, RGB color, and depth imagery and then deploy already developed deep learning models to analyze the input data in real time. The embedded system analyzes and returns the processed images via wireless streaming, where they can be viewed remotely and relayed back to the firefighter using an augmented reality platform that visualizes the results of the analyzed inputs and draws the firefighter's attention to objects of interest, such as doors and windows otherwise invisible through smoke and flames.

\end{abstract}


\section{Introduction}

\IEEEPARstart{F}{irefighting} is an inherently dangerous and potentially fatal activity. Even the most experienced firefighters can be affected by stress and anxiety during a fire, leading to disorientation and the corresponding impairment of situational awareness. Furthermore, these stressors are only exacerbated by the smoke, high temperatures, and low visibility at the scene, further compromising a firefighter’s ability to effectively respond to the situation. Situational awareness, or the real-time knowledge of continuously changing conditions at the scene, is essential to accurate decision-making and the ability of a firefighter to maintain situational awareness in these conditions is likewise critical. The firefighter’s life and the lives of those needing rescue are reliant on the firefighter's ability to make accurate decisions during scene navigation, and recent advancements in deep learning \cite{lecun2015deep}, data processing, and embedded systems technologies now offer a new way to assist and improve this decision-making process.

Indeed, such technologies are already being implemented in firefighting applications. Just within the last decade, these applications have grown to include the use of wireless sensor networks (WSN) for early fire detection and response \cite{wilson2007wireless}, virtual reality platforms for firefighter training \cite{vichitvejpaisal2016firefighting, yuan2012building, yang2018train}, and autonomous or semi-autonomous firefighting robots \cite{lawson2016touch, ranaweera2018shortest,liu2016robot}. But fighting fires poses many challenges to the firefighters themselves that are not adequately addressed by these solutions. Continuously changing life threatening situations can engender severe stress and anxiety, and induce disorientation that can lead to inaccurate decision making and decreased situational awareness. Recent research \cite{li2014situational} has demonstrated a discrepancy between the information available during fire emergencies and the information required for critical situational awareness, as well as the importance of information accuracy to error-free decision making. 

In fact, inaccurate judgment calls are consistently among the top causes of firefighter fireground injuries each year, and firefighter personal protective equipment (PPE), contrary to its purpose, is often a contributing factor \cite{petrucci2016inaccuracy}. Although firefighter PPE is designed to protect the wearer during extreme fire conditions, it is not without its drawbacks, including reduced visibility and range of motion \cite{petrucci2016inaccuracy, kozlovszky2014environment}. These drawbacks have led to the integration of various sensors, such as infrared cameras, gas sensors, and microphones, into firefighter PPE to assist and monitor firefighters remotely \cite{kozlovszky2014environment}. The data acquired by these sensors can also be leveraged for an additional purpose: to provide firefighters with object recognition and tracking and 3D scene reconstruction in real-time, creating a next generation smart and connected firefighting system.

\begin{figure*}
    \centering
    \includegraphics[scale=0.7]{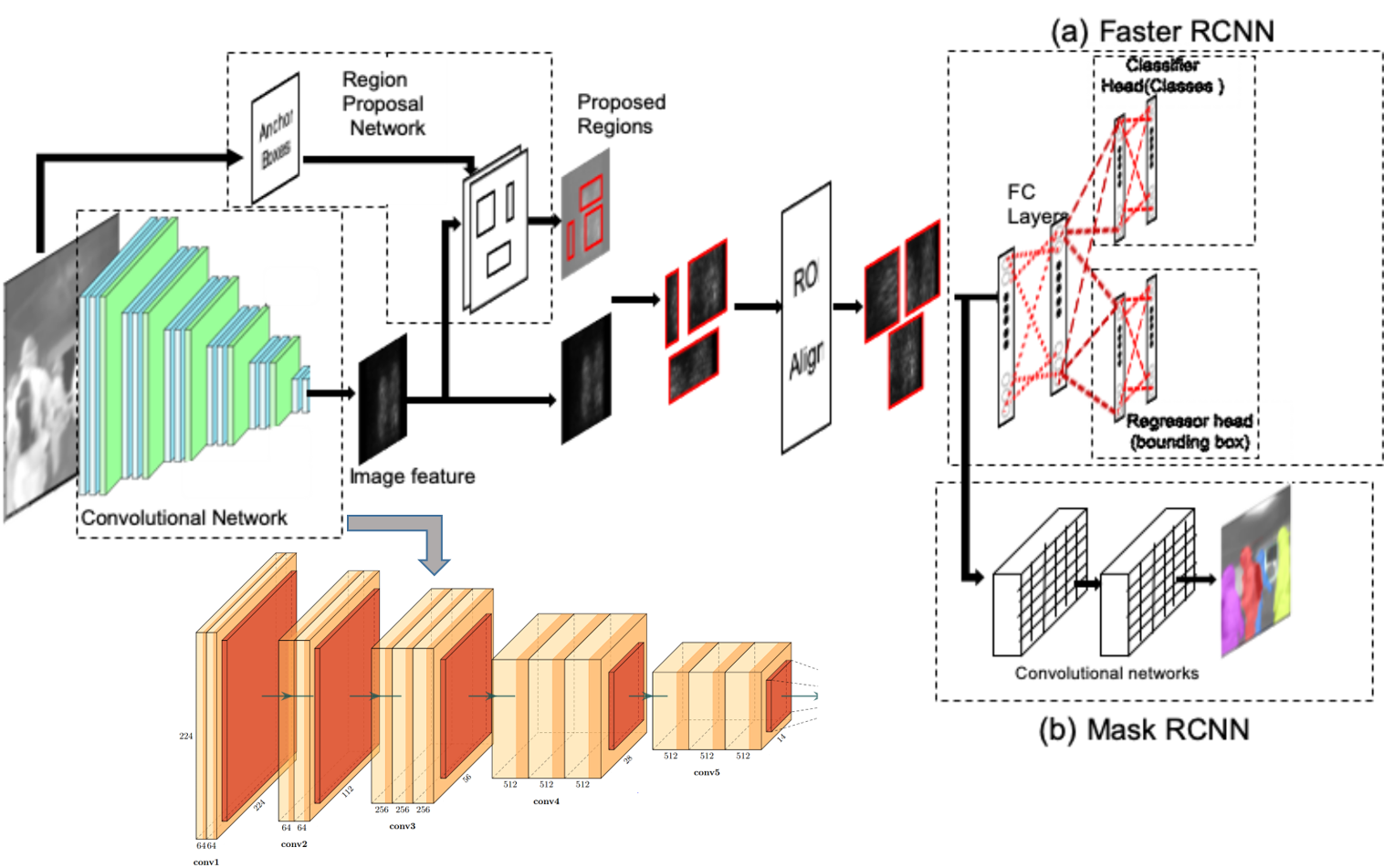}
    \caption{Object detection and segmentation framework. Both share a common Region Proposal Network(RPN) for significant feature extraction. For object detection/tracking, the  processed output of the RPN is fed to a classifier and a regressor to  output class scores and bounding box co-ordinates respectively. This is performed by Faster RCNN as shown in (a). For image segmentation, the same processed output of RPN is fed to a Convolution Network that generates a mask when over-layed with the original image, resulting in semantic segmentation as its output (b).}
    \label{fig:structure}
\end{figure*}
The procedure for object and human detection can be used to describe
the scene by using human language, prioritize the presentation of images to the
commander depending on the contents of the image, or to find different scenarios
on demand (by presenting those scenes that contain fire, procumbent humans, or other
situations of interest). Another interesting application is to produce augmented
reality for the fire fighter in the scene. The situations of interest where this implementation can be applied are diverse. For instance, fire fighters may find themselves in a scenario where their vision is impaired due to dense smoke or dust particles, or where there are no visible light, so the scene has to be illuminated with a handheld or helmet flashlight. In general, in these situations, the thermal cameras can assist in improving the visual clarity of the scenario. Nevertheless, thermal images are more difficult to interpret than visible light-based images. A way to improve the perceptual experience
of the fire fighters is to use the detected images to reconstruct the scenario using augmented reality glasses. We have developed preliminary prototypes of an
augmented reality system that projects the images detected and segmented by a
neural network to the Hololens augmented reality glasses, commercialized by Microsoft. They project the image from a computer
over a semitransparent screen, so the user can see the projected images, and also through them. The images are
projected so their virtual focal point is about 10 inches away from the user’s eyes,
thus producing the illusion of seeing a semitransparent screen floating over the
user. Their see-through effect allows the user to interact with the environment without being impeded or disoriented by a virtual experience.
The headset has an incorporated inertial measurement unit (IMU) that is used to
determine the position of the head. This information is useful to maintain the position of the image when the user rotates their head.

The aim of this paper  is to demonstrate the hardware deployment of the firefighter aid system. This proposed system  integrates FLIR One G2 and Intel RealSense D435i thermal and depth cameras with an NVIDIA Jetson embedded Graphics Processing Unit (GPU) platform that deploys already developed machine and deep learning models \cite{bhattarai2020deep} to process captured datasets and then wirelessly stream the augmented images in real time to a secure local network router based on a mesh node communication topology \cite{hamke2019mesh}. The augmented images are then relayed to a Microsoft HoloLens augmented reality platform incorporated into a firefighter's PPE as part of a system that is insusceptible to environmental stressors while also being able to detect objects that can affect safe navigation through a fire environment with a greater than 98\% test accuracy rate \cite{bhattarai2020deep}. This relay, interpretation process and return of interpreted and projected data is real-time. 

\section{The Deep CNN Model}
This research deploys a trained deep Convolutional Neural Network (CNN) based autonomous system to identify and track objects of interest from thermal images captured at the fireground.
While most applications of CNN systems to date are found in the related fields of surveillance and defense, some recent work has explored applications in unmanned aerial vehicles (UAVs) to detect pedestrians \cite{de2018using} and avalanche victims \cite{rodin2018object} from the air. Other recent work has demonstrated the use of CNN systems on embedded NVIDIA Jetson platforms, such as the TX1 and TX2, for traffic sign identification (with applications for autonomous vehicles) \cite{han2017traffic} and autonomous image enhancement \cite{nokovic2019image}. Other work has deployed fully-Convolutional Neural Networks (FCNNs) on the TX2 to provide single-camera  real-time, vision-based depth reconstruction \cite{bokovoy2019real} and demonstrated an attention-based CNN model that can learn to identify and describe image content that can then be generated as image captions \cite{xu2015show}. 

\begin{figure*}[!t]
\centering
\includegraphics[width=0.8\linewidth]{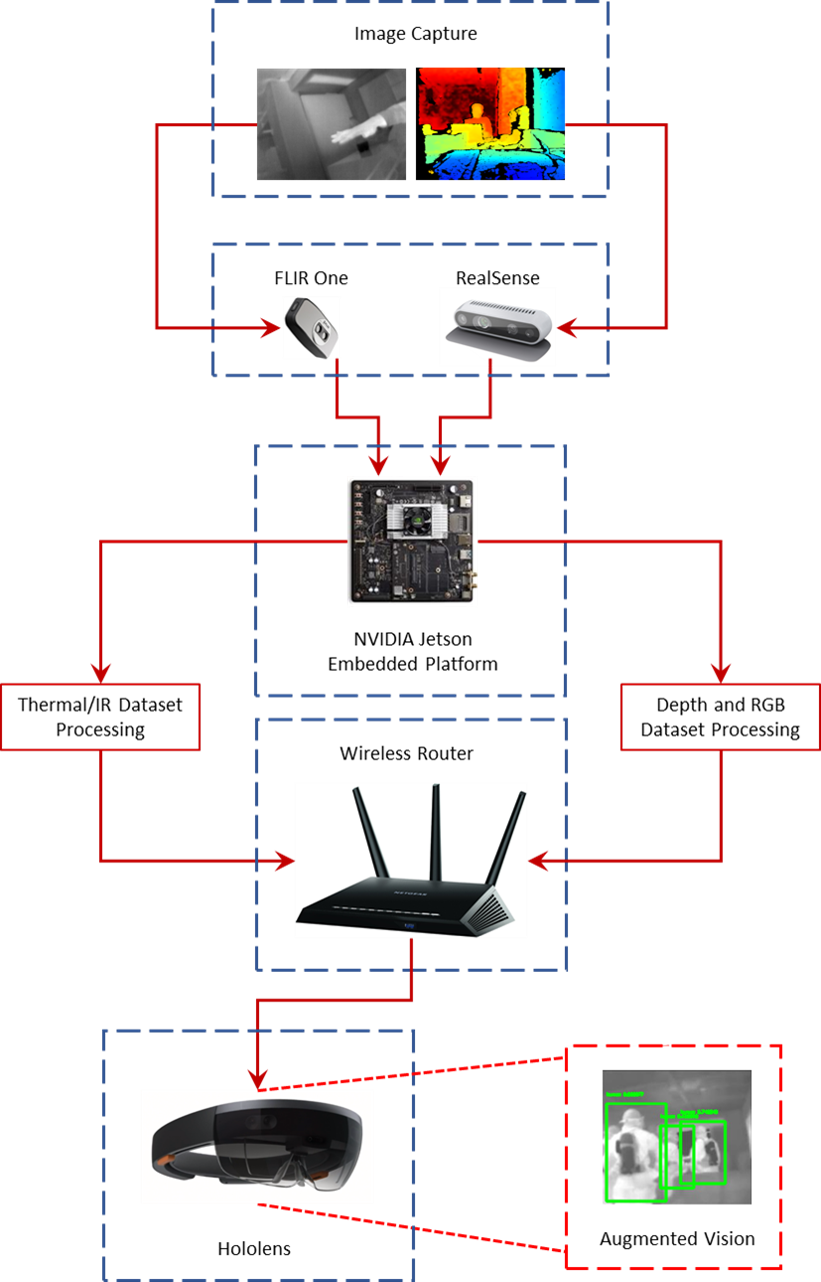}
\caption{Overview of relationships between system components}
\label{overview}
\end{figure*}

The embedded system described in this paper deploys a deep CNN system, but is a novel approach that, to the best of our knowledge, is the first of its kind \cite{bhattarai2020deep}. In order to implement the augmented reality system, we constructed a light weight Mask RCNN based  neural network as shown in Figure~\ref{fig:structure}.This framework is built on top of a VGG-16- like model.  To accomplish the object segmentation and tracking tasks, we extended the VGG model framework with proposal networks to construct a  region-based CNNs (RCNNs). These models are able to localize each one of the objects in the scene and continuously track them. The model was further modified with the addition of a 1x1 Convolutional layer constructing a Mask RCNN to achieve better detection and to produce, at the same time, a semantic segmentation of the scene. The system was trained with a large quantity of images recorded in real fire training situations by firefighters to identify and track objects in real time \cite{bhattarai2020deep}. 

Objects are identified and tracked in images from the thermal imaging dataset using Faster RCNN \cite{girshick2015fast}, and then instances of significant objects are masked and shaded using Mask RCNN \cite{he2017mask}.  While the Faster RCNN uses a fully connected neural network for classification and regression for bounding box coordinates estimation, the Mask RCNN uses a set of $1\times 1$ convolutional layers to estimate the segmentation of each one of the objects. The previous layers in both architectures are identical in structure. Therefore, in our approach we construct a hybrid network containing the outputs of both networks in a single structure. The combination of both outputs produces a sequence with the objects in the scene segmented and classified. These algorithms effectively exploit the images gathered from the infrared camera by using a trained deep CNN system to identify, track, and segment objects of interest.
The tasks of object recognition and tracking have the structure depicted in Figure \ref{fig:structure}.  \par
Furthermore, this system is capable of performing human recognition and posture detection to deduce a victim’s condition and guide firefighters accordingly, which can assist in prioritizing victims by urgency. Processed information can be relayed back to the firefighter via an augmented reality platform that enables them to visualize the results of the analyzed inputs and draws their attention to objects of interest, such as doors and windows otherwise invisible through smoke and flames. This visualization also provides localized information related to those objects.

\section{The Embedded System}

\subsection{System Overview}
The embedded system consists of the following commercial off-the-shelf (COTS) hardware components:
\vspace{\baselineskip}
\begin{itemize}
\item{NVIDIA Jetson TX2 Development Kit}
\item{FLIR One G2 for Android thermal camera}
\item{Intel RealSense D435i depth camera}
\item{Wireless network router}
\end{itemize}
\vspace{\baselineskip}
An overview of the interactions between these components and their place within the smart and connected firefighting system is illustrated in Figure~\ref{overview}. The components can be further categorized into those that acquire data (FLIR One and RealSense cameras), those that process data (NVIDIA Jetson), those that communicate data (NVIDIA Jetson and wireless router), and those that ultimately augment data (HoloLens). Each of these categories is described here in detail.

\begin{figure}[t]
\centering
\includegraphics[width=3.49in]{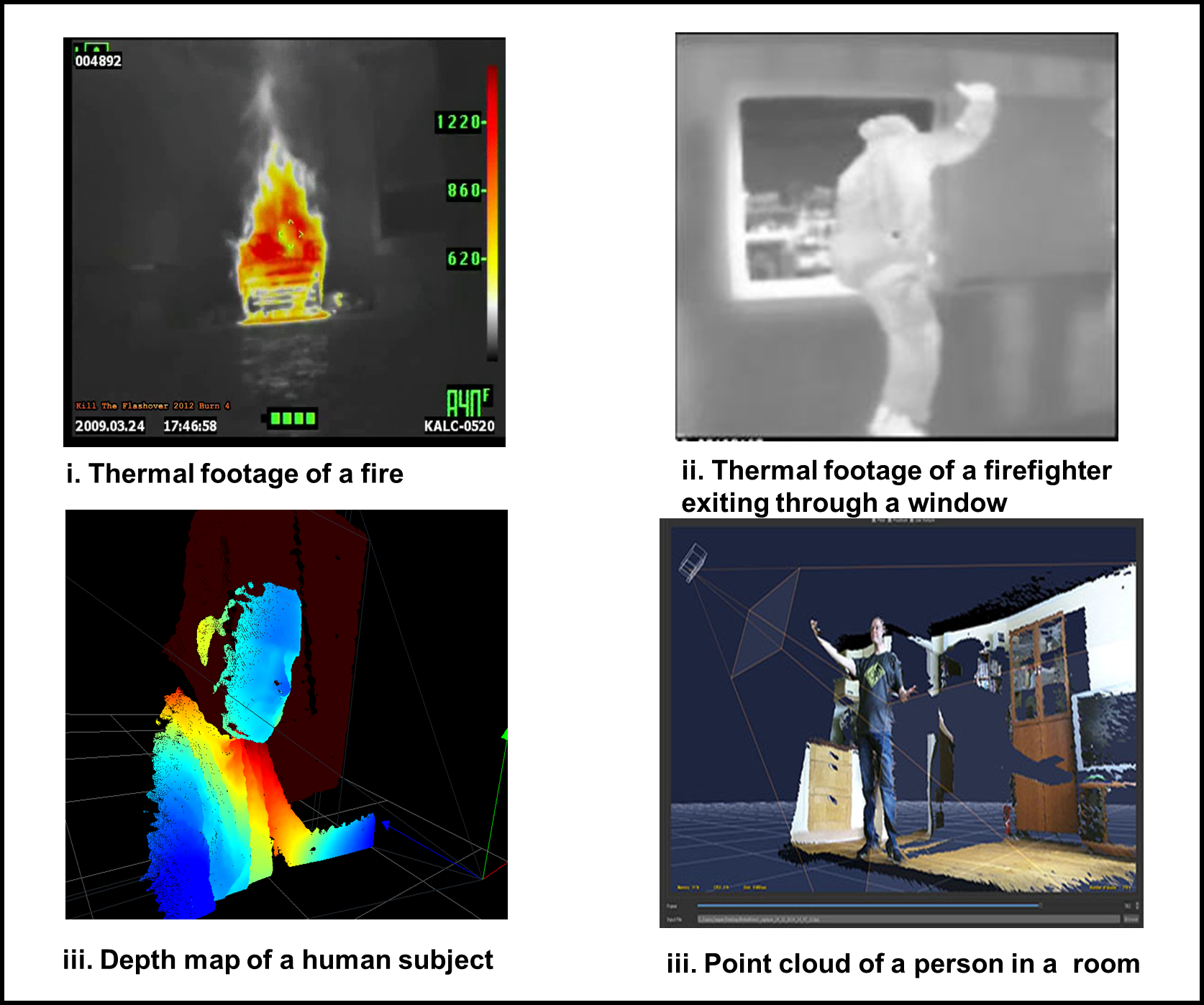}
\caption{Types of datasets acquired by different cameras}
\label{datasets}
\end{figure}

\subsection{System Architecture}
\subsubsection{Data Acquisition}
The system acquires data utilizing the live streaming of thermal, RGB, and depth image datasets captured by both the FLIR One and Intel RealSense cameras. Examples of such datasets are shown  in Figure~\ref{datasets}. Each camera has the option to adjust frame rate, but for the purposes of consistency, all tests were conducted at a rate of 30 frames per second (fps). Both cameras are attached to Jetson platforms via USB cables, and the acquired datasets are transmitted from the camera sensors to the embedded Jetson platforms through the same USB connections.

\subsubsection{Data Processing}
The camera-acquired datasets, which include thermal, RGB, and depth images, are processed by the deep learning models deployed on the Jetson GPU platforms. Each captured frame is processed by the algorithms and the rendered images are then streamed to a secure wireless network router, along with the original raw images. In addition, the images are also saved locally on the Jetson's onboard flash memory;available storage on the embedded platform is limited however, and stored images are only stored until a limit is reached (up to 100 captured frames), at which point the saved frames are removed and a new stored set is begins.

The specific information processed by the deep learning model is object detection, object tracking, instance segmentation, and scene description.

\subsubsection{Data Communication}
The processed and raw images are wrapped in HTML code and made available for access by connected devices on the secure local network through a wireless router as a video stream. This allows access to both the firefighter whose sensors captured the data and also to commanding officers overseeing and guiding the response effort as well as any other firefighters at the scene.

\subsubsection{Data Augmentation}
The processed data is relayed in real time through the network for visualization on the Microsoft HoloLens, thus providing the firefighter with the visualized augmented information to assist navigation and enhance situational awareness.

\subsection{System Implementation}
\subsubsection*{Camera Sensor Implementation}
Both the FLIR One and the RealSense D435i cameras are available commercially; however, neither camera was designed for the ARM64 architecture of the NVIDIA Jetson platforms, which run a version of embedded Ubuntu Linux OS developed specifically for the Jetson's Tegra processor (a system on a chip, or SoC). Therefore, in order for the system to recognize and capture data from the USB connected cameras, it was first necessary to build each camera its own kernel driver module for the Tegra Ubuntu kernel. These modules are variations of the open source v4l2loopback Ubuntu Linux kernel module, modified for the individual camera.

\begin{figure}
    \begin{subfigure}[b]{0.5\linewidth}
        \centering\includegraphics[width=\linewidth]{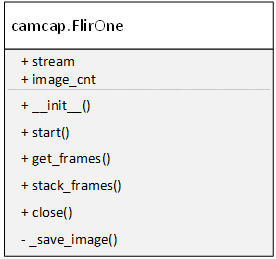}
        \caption{FlirOne class}\label{fig:flir_uml}
    \end{subfigure}%
    \begin{subfigure}[b]{0.5\linewidth}
        \centering\includegraphics[width=\linewidth]{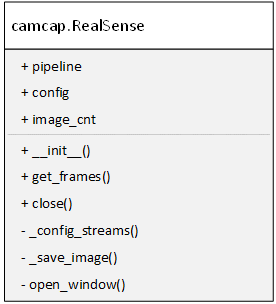}
        \caption{RealSense class}\label{fig:rs_uml}
    \end{subfigure}
\caption{UML class diagrams}\label{fig:uml}
\end{figure}
\begin{figure}[!t]
\centering
\includegraphics[width=3.49in]{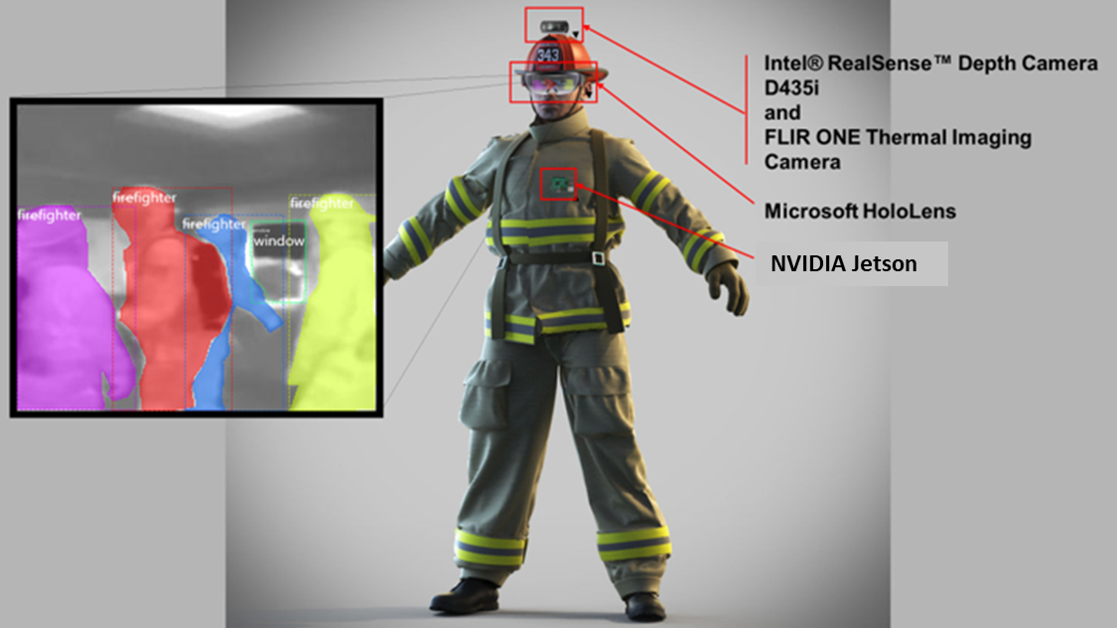}
\caption{Embedded system deployment}
\label{deployment}
\end{figure}

The cameras are controlled at the user level by the Python language wrapper programs, flircap.py and rscap.py, for the FLIR One and Realsense cameras respectively. These programs implement the FLIR One and RealSense classes described in the camcap.py interface. Each class provides the necessary user-level controls to initialize the cameras and acquire and save frames, as described in Figure~\ref{fig:uml}

The individual camera wrapper programs, flircap.py and rscap.py, initialize the cameras and implement OpenCV to process the individual captured frames with the trained deep CNN system using the Jetson's onboard GPU.

\begin{figure}[!t]
\centering
\includegraphics[width=3.49in]{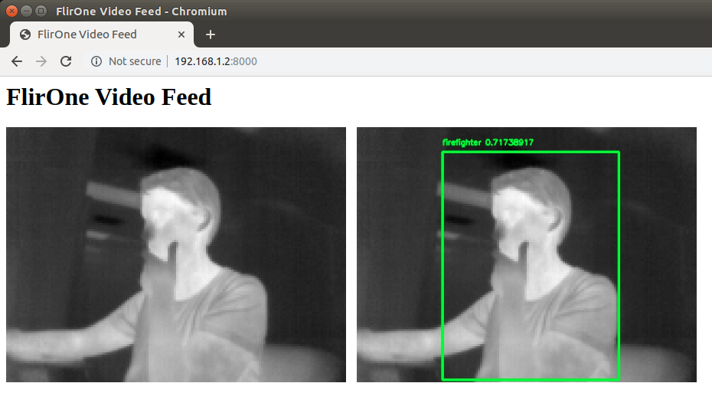}
\caption{FlirOne video feed raw and processed output}
\label{fig:flir_feed}
\end{figure}

\begin{figure}[!t]
\centering
\includegraphics[width=3.49in]{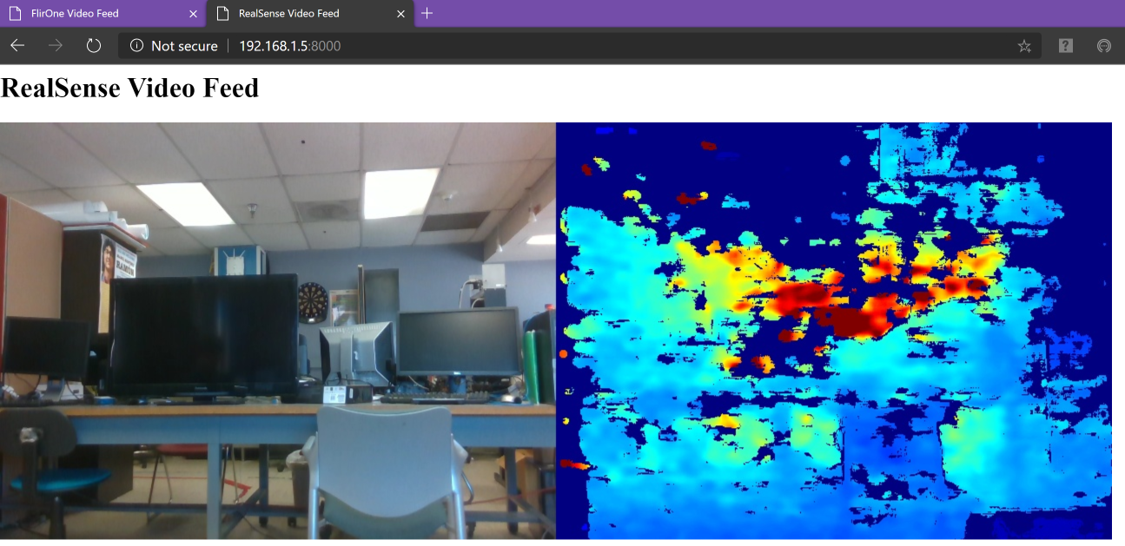}
\caption{RealSense RGB and depth colormap video feed output}
\label{fig:rs_feed}
\end{figure}
\subsection{System Deployment}

The data acquisition, data processing, and augmented data hardware components of the embedded system seen in Figure ~\ref{overview} will be integrated with the firefighter's personal protective equipment (PPE) as shown in Figure \ref{deployment} to provide the firefighter the real time information processing and augmentation that will enhance situational awareness and assist in minimizing the number of poor decisions made due to stress, anxiety, and disorientation. The system is lightweight and can easily be integrated with the firefighter suit. 

\section{Discussion and Results}
\begin{figure*}
\centering
\begin{subfigure}[b] {0.9\textwidth}
   \includegraphics[width=0.9\linewidth]{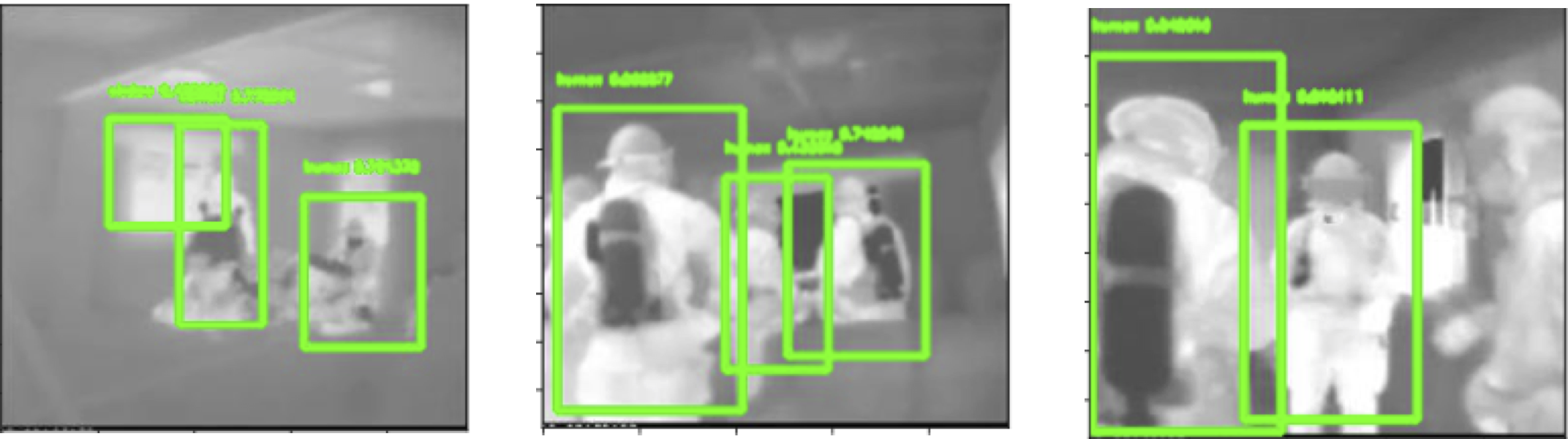}
   \caption{Demonstration of  object detection and tracking implemented with Faster RCNN. The green boxes bind the object of interest. The label on top infers corresponding object and detection probability. Selected objects of interest include firefighters and civilians, doors, windows, and ladders.}
   \label{fig:FasterRCNNTracking} 
\end{subfigure}

\begin{subfigure}[b]{0.9\textwidth}
   \includegraphics[width=0.9\linewidth]{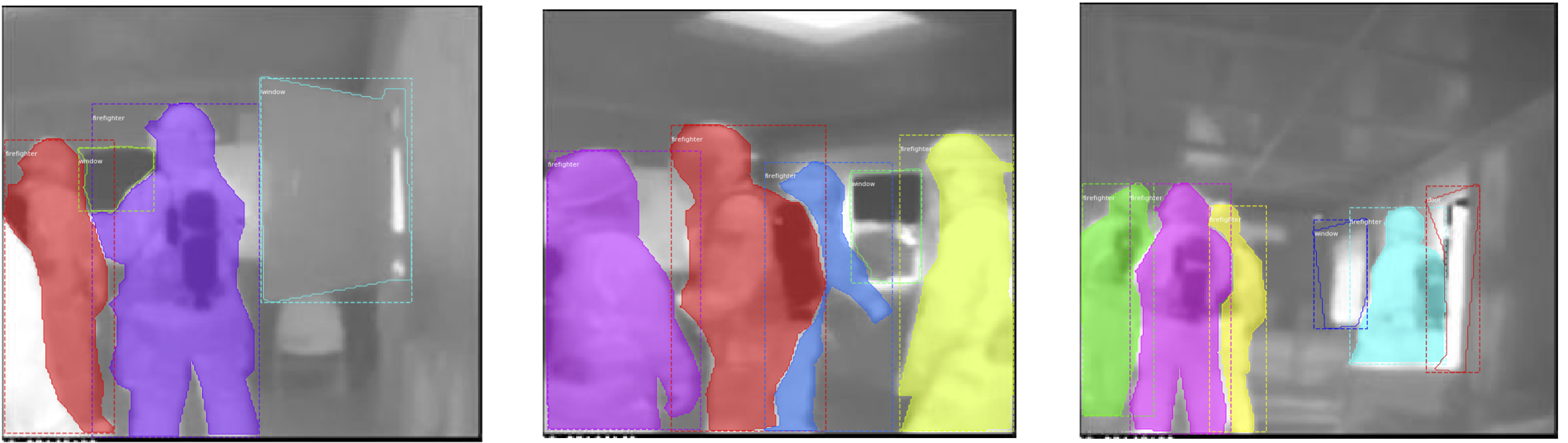}
   \caption{Demonstration of  object instance segmentation implemented with Mask RCNN. Every instance of significant object is enclosed with different mask . The instances for firefighter are shaded with distinct colors. Doors and windows are shaded with instance boundaries. }
   \label{fig:MaskRCNNTracking}
\end{subfigure}
\caption{Results of the proposed framework } \label{fig:Tracking}
\end{figure*}

 The resulting live stream output of flirap.py (both raw and processed thermal images) is illustrated in Figure~\ref{fig:flir_feed}. Similarly, the live stream output of rscap.py (the RGB color and color mapped depth images captured by the RealSense camera) is illustrated in Figure~\ref{fig:rs_feed}. The results of the algorithm of the thermal captured feeds in firefighting scenarios is shown in Figure~\ref{fig:FasterRCNNTracking} and Figure~\ref{fig:MaskRCNNTracking}.
 
 Figure \ref{fig:FasterRCNNTracking}  shows the result of the object detection and tracking, where each object
is confined in a bounding box with a detection probability. This is performed by the Faster RCNN described in section II. By using our modified R-CNNs, the objects
are segmented in different colors and tracked in real time as shown in Figure~\ref{fig:MaskRCNNTracking}. The sequences of segmented images are represented in the Hololens. The raw depth maps and RGB feeds are presented in \ref{fig:rs_feed}. Currently, these two different feeds are combined to construct the 3D map of the scene via Intel Realsense built in a 3D reconstruction algorithm. We are currently working to combine these feeds with the thermal feed to construct a better 3D map in low light conditions.  \par

The work presented here lays a foundation for the development of a real-time situational map of the structural configuration of a building that is actively built and updated via the live thermal imagery being recorded by firefighters moving through the scene. An initial demonstration of indoor positioning and path planning is presented in \cite{vadlamani2020novel} which is based on the estimation of camera movement through estimation of the relative orientation with SIFT and Optical flow.  This map, which is updated in real-time, could be used by firefighters to assist them in safely navigating the burning structure and improve the situational awareness necessary in decision making by tracking exits that may become blocked and finding alternatives. Utilizing the features detected via the approach presented in this paper, a robust localization and tracking system to track objects of interest in sequences of frames can be built. The visual features from this framework have also been coupled with a Natural Language Processing (NLP) system for scene description and allow the framework to autonomously make human-understandable descriptions of the environment to aid firefighters to improve their understanding of the immediate surroundings and assist them when anxiety levels are heightened. Our future work seeks to join these two components with a reinforcement learning (Q-learning) algorithm that utilizes the continuously updated state map and reinforcement learning techniques that assist in path planning and can be vocalized through the NLP system. The deep Q-learning based approach provides a navigation system that actively avoids hazardous paths. These three components, built on the backbone of the research presented here, can be fused to accomplish the ultimate goal of providing an artificially intelligent solution capable of guiding firefighters to safety in worst-case scenarios. Further to compress the neural net parameters into an embedded platform for a real time performance, we would like to explore the tensor train networks\cite{novikov2015tensorizing} and distributed implementation \cite{bhattarai2020distributed} for large scale data.  In moments of panic, our system may assist fire fighters to regain a sense of their surroundings and the best path they can take to exit the burning structure.

\subsection{Conclusion}
We have created an automated system that is capable of real-time object detection and recognition using data gathered on site. The algorithms developed are able to provide detailed analyses of the surroundings. The improvements made to the firefighter’s ability to maintain a constant and consistent situational awareness also greatly improves their ability to correctly interpret surroundings, maximizing rescue efficiency and effectiveness.

The embedded system is capable of initializing and capturing real-time thermal images from a FLIR One G2 camera, processing those images for object detection, object tracking, instance segmentation, and scene description on an NVIDIA Jetson TX2 embedded GPU development platform, and returning the output as a real-time live stream over a secure wireless network to first responders and commanding officers alike. In addition, the embedded system is also able to acquire and process RGB color and depth images captured by an Intel RealSense D435i camera on an NVIDIA Jetson AGX Xavier embedded GPU development platform and to return the output as a second real-time live stream over the same secure wireless network.

There are a number of challenges to the successful implementation of a smart and connected firefighter system, and primary among them is the challenge of providing robust, real-time data acquisition, processing, communication, and augmentation. This research seeks a foundation on which solutions to these challenges can be overcomed.

Our next steps will focus on fusing the data collected from the different cameras to form a 3D map of the building. The data will form the foundation of a new AI system that assists the firefighters with path planning through cataloging of doors, windows, and paths clear of fire. We will also address the hardware issues  which currently limit the application of the embedded system at the fireground.


\section*{Acknowledgment}
This work was funded by NSF grant 1637092. We would like to thank the UNM Center for Advanced Research Computing, supported in part by the National Science Foundation, for providing the high performance computing, large-scale storage, and visualization resources used in this work. We would also like to thank Sophia Thompson for her valuable suggestions and contributions to the edits of the final drafts.

\bibliographystyle{IEEEtran}
\bibliography{IEEEabrv,references}

\end{document}